\documentclass[letterpaper, 10 pt, conference]{ieeeconf}  
\pdfoutput=1

\IEEEoverridecommandlockouts 
\usepackage{hyperref}
\usepackage{float}
\usepackage[utf8]{inputenc}
\usepackage{graphicx}
\usepackage{subcaption}
\usepackage{algorithmic}
\usepackage[ruled,vlined]{algorithm2e}
\usepackage{siunitx}
\usepackage{amsmath}
\usepackage{tikz}
\usepackage{pgfplots}
\usepgfplotslibrary{dateplot}
\usepackage{adjustbox}
\usepackage{tabularx,ragged2e,booktabs,caption}
\usepackage{comment}
\newcolumntype{L}{>{\RaggedRight\arraybackslash}X}
\newcommand{\COMMENT}[2][.5\linewidth]{
  \leavevmode\hfill\makebox[#1][l]{//~#2}}

\usepackage[backend=biber,style=numeric,sorting=none]{biblatex}
\title{\LARGE \bf
Robotics During a Pandemic: The 2020 NSF CPS Virtual Challenge -- SoilScope, Mars Edition}

\author{Darwin Mick, K. Srikar Siddarth, Swastik Nandan, Harish Anand, Stephen A. Rees, Jnaneshwar Das
{\small
\thanks{Authors are with the School of Earth and Space Exploration, Tempe, Arizona {dpmick, snandan3, hanand4, jdas5@asu.edu}, Stephen is a staff at Vanderbilt University, Nashville, Tennessee {stephen.a.rees@vanderbilt.edu}, and Srikar is with the  Dept. of ECE, National Institute of Technology Karnataka, India {kodihallisrikarsiddarth.181ec218@nitk.edu.in} }%
}}

\begin{document}

\maketitle
\thispagestyle{empty}
\pagestyle{empty}
\begin{abstract}

Remote sample recovery is a rapidly evolving application of Small Unmanned Aircraft Systems (sUAS) for planetary sciences and space exploration. Development of cyber-physical systems (CPS) for autonomous deployment and recovery of sensor probes for sample caching is already in progress with NASA's MARS 2020 mission. To challenge student teams to develop autonomy for sample recovery settings, the 2020 NSF CPS Challenge was positioned around the launch of the MARS 2020 rover and sUAS duo. This paper discusses perception and trajectory planning for sample recovery by sUAS in a simulation environment. Out of a total of five teams that participated, the results of the top two teams have been discussed. The OpenUAV cloud simulation framework deployed on the Cyber-Physical Systems Virtual Organization (CPS-VO) allowed the teams to work remotely over a month during the COVID-19 pandemic to develop and simulate autonomous exploration algorithms. Remote simulation enabled teams across the globe to collaborate in experiments. The two teams approached the task of probe search, probe recovery, and landing on a moving target differently. This paper is a summary of teams' insights and lessons learned, as they chose from a wide range of perception sensors and algorithms. 

\href{https://cps-vo.org/group/CPSchallenge}{Web page: https://cps-vo.org/group/CPSchallenge}
\end{abstract}

\section{Introduction}
\label{sec:intro}


Small Unpiloted Aircraft Systems (sUAS) are gaining popularity in agricultural and environmental applications. sUAS are used for vegetation mapping, crop stress monitoring, and fruit counting through a suite of onboard sensors \cite{dasCASE2015, das2016}. Furthermore, sUAS have been effectively used in missions to retrieve physical samples for ex-situ analysis. For instance, they have been used for the aerial collection of water samples and aerobiological sampling in agricultural fields ~\cite{ore2015, schmale2008development}. 

The NSF-sponsored 2020 CPS Challenge is motivated by the plethora of applications of sUAS \cite{cpsc1, cps}. The original intention was to create an entirely autonomous system with sUAS to deploy and retrieve a soil probe that could measure the Sonoran Desert's various soil properties. However, during the COVID-19 pandemic, building and testing of the hardware became infeasible. The competition moved to a virtual environment to adapt to these new challenges and changed its scope to follow the Mars 2020 theme. 

OpenUAV, a web-based robotics simulation software that enables students and researchers to run robotic simulations on the cloud, provided easy access to sUAS algorithm development and experiments, despite the pandemic \cite{anand2019openuav}. It also allowed students from various countries to interact and compete without the time-consuming process of setting up the software and hardware themselves, which is typical in a robotics competition. The need for a remote testbed has already been recognized and implemented in a physical setting \cite{fink2011robotic}. However, OpenUAV is completely virtual and effective without access to hardware. 

\begin{figure}[htpb]
    \centering
    \includegraphics[width=3in]{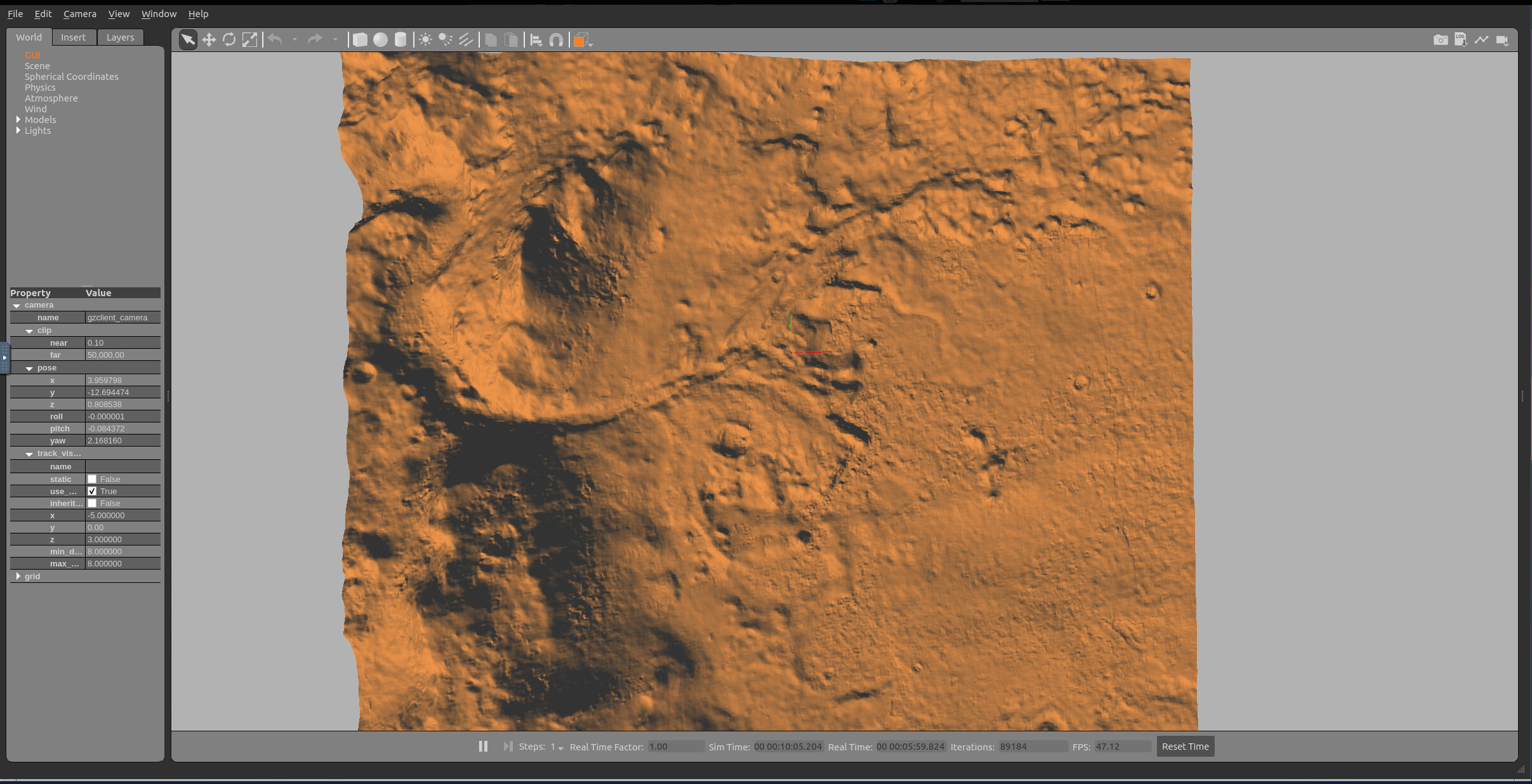}
    \caption{ {\small}The Martian Jezero Crater competition site shown in the Gazebo simulator within the OpenUAV CPS-VO framework. }
    \label{fig:jezero_mesh}
\end{figure}

Inspired by the Mars Perseverance Rover and the Mars Helicopter Ingenuity, the new challenge involves a simulated Martian environment, where teams seek to autonomously retrieve and deploy a soil probe using an sUAS and then return the sUAS to a moving rover. Studies have already been conducted showing the efficacy of operating an sUAS on the surface of Mars for scouting purposes, and Ingenuity will test them in practice this year~\cite{grip2017flight,balaram2018mars}. Furthermore, retrieving and testing a diverse set of Mars soil samples is one of Perseverance's main science goals~\cite{mustard2013report}. Therefore, soil sampling and retrieval are of interest to the scientific community, and using an sUAS to retrieve samples is not only feasible but could also allow for samples to be retrieved from locations that are hazardous for rover traversal. A future Mars mission where an sUAS is responsible for critical path science goals, such as probe retrieval, is highly likely, and this challenge lays the groundwork for three autonomy solutions to this specific problem:

\begin{itemize}
\item Autonomously navigate to a search location to search for, detect, descend upon, and pick up the soil probe.
\item Autonomously navigate to a drop location, descend, and detach the soil probe.
\item Autonomously return to and land on a moving rover.
\end{itemize}


In this work, we present two sUAS methods that can autonomously retrieve and deploy a lightweight probe, then land on a moving rover. We also discuss the unique challenges of hosting a competition during a pandemic and how it affected the participating teams.

This paper follows the methodologies of two different teams, and the structure is as follows. In section~\ref{sec:Competition_Description}, we describe the simulation environment, followed by a description of our solutions in section ~\ref{sec:method}. We discuss simulation results in section~\ref{sec:experiments}, followed by a discussion of the pandemic's impact on the results in section~\ref{sec:discussion}. Finally, conclusions and future work are in section~\ref{sec:future-work}. 

\section{Competition Description}

\label{sec:Competition_Description}

\subsection{Competition Phases}
\subsubsection{Phase 1}The first phase of the competition was structured for two weeks and targeted visual servoing of an sUAS with a sensor probe for autonomous recovery. This was followed by autonomous deployment of the probe at a specified location. For this phase, a terrain mapped with a multi-rotor sUAS with an onboard MicaSense RED Edge MX camera was used. The 3D mesh was inserted in all OpenUAV container environments, along with a custom sUAS model and sensor probe.

\subsubsection{Phase 2}
The second phase of the competition built upon Phase 1 and involved a significantly larger site based on a Jezero Crater MARS 2020 site mesh model acquired in Spring 2020. This phase requires the sUAS to land on the trunk of the rover while the rover is in motion.
\label{sec:sys-desc}

\subsection{System Description}

\begin{figure}[htpb]
\centering
\includegraphics[width=3in]{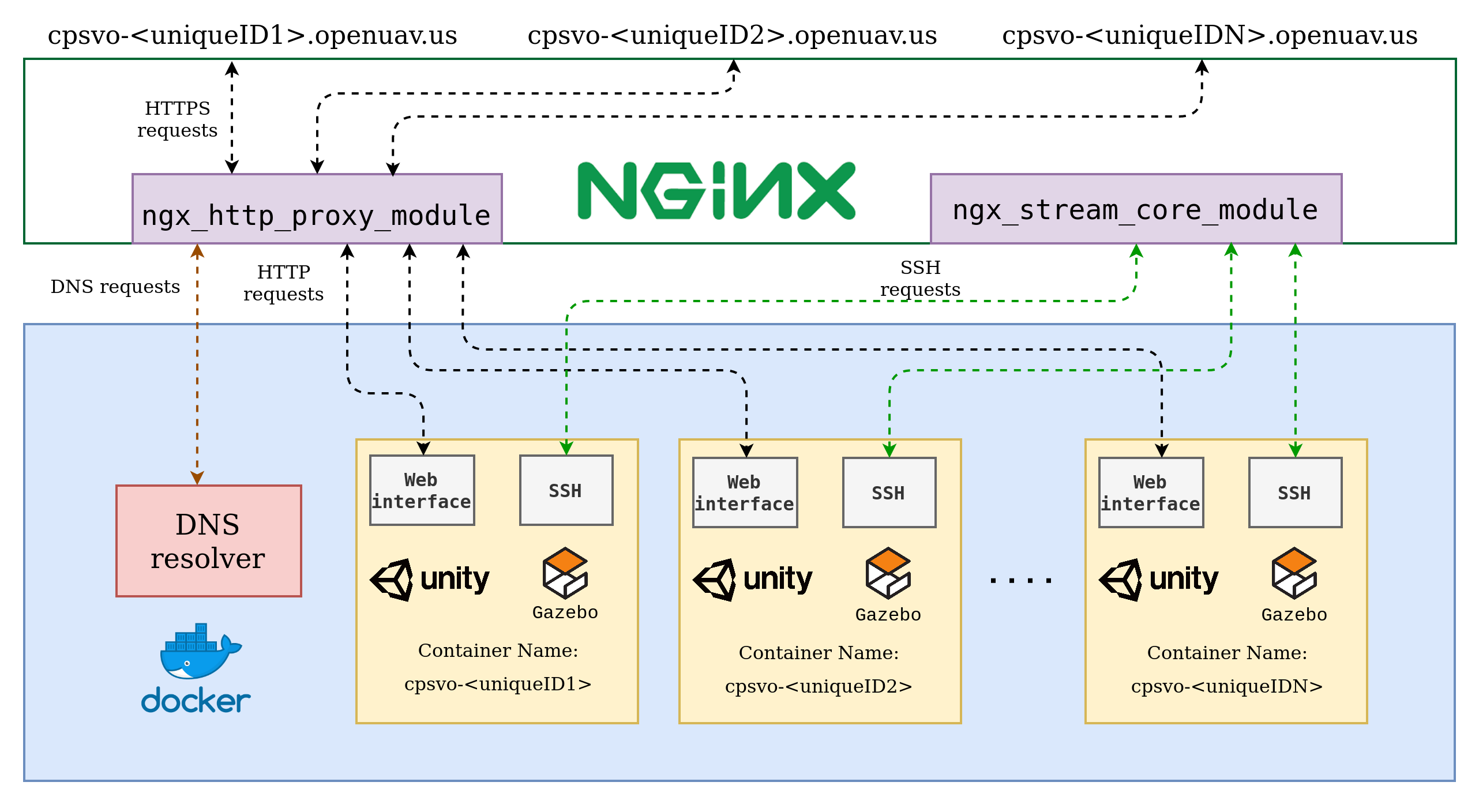}
\caption{\label{fig:uav-booth-orange} {\small} The architecture of OpenUAV involves a dockerized Lubuntu desktop environment containing robotics software like Gazebo, Unity, ROS, and PX4. Nginx redirects the web requests to the corresponding docker container.
}
\end{figure}

A combination of two 2.1 GHz 16 core processors, two NVIDIA GeForce RTX 2080ti, 2 TB SSD, and 10 TB HDD forms our server's hardware framework. A total of 107 containers were created during the CPS Challenge period \cite{anand2019openuav}. A simulation session consists of a Lubuntu desktop equipped with widely used robotics software, that is accessible through a web URL. By utilizing VirtualGL, any 3D rendering calls by the applications directly go to the GPU, and hence, the rendering looks smoother. Fig. \ref{fig:uav-booth-orange} shows the OpenUAV platform's architecture that enabled a virtual competition during the pandemic. Further details on the architecture and implementation can be found on the OpenUAV paper \cite{anand2019openuav}.

\subsection{Software}
\subsubsection{ROS and Gazebo}
ROS is a framework used to easily communicate between the robot and ROS nodes through ROS messages \cite{quigley2009ros}. Gazebo is a simulator that comes with a physics engine \cite{koenig2004design}. Together, they make the basic framework of the robotic simulation. The MAVlink communication protocol allows communication between the UAV model in the simulation with the code written by the user.

\subsubsection{PX4, MAVROS, and QGC}
PX4 is an open-source flight control software for aerial, ground, and sea vehicles \cite{meier2015px4}. PX4 enables low-level rate control and higher-level velocity and position control of the UAV. It also provides sensor support, flight logging, and the state estimator. QGroundControl is a remote flight planning and tracking software that enables the teams to control the vehicles as they would normally do in an outdoor field experiment. MAVROS is a ROS package that acts as the liaison between PX4, QGC, and ROS. 

Using software in the loop (SITL), PX4 and MAVROS reduce the gap between simulation and real testing. The software controlling the real sUAS runs in two parts, the higher-level functionalities run on an Intel NUC and the flight controller on the Pixhawk Autopilot. During the simulation, the PX4 SITL takes the role of the Pixhawk Autopilot.  

\subsection{Reproducibilty}
All participant's code and container environments are available on the CPS-VO page, where any future teams and experts can redo the experiments and improve upon the participant's submissions \cite{cps}.


\section{Methodology}
\label{sec:method}

The purpose of the NSF CPS 2020 challenge has been to let contestants explore from a range of perception sensors and develop innovative strategies for trajectory planning and state estimation. State estimation of sUAS is a unique challenge due to the aircraft's persistent pitch angle while searching for probes. Moreover, the jerks due to aggressive change of modes result in noisy state estimation. A robust strategy is needed to incorporate the inaccuracy in the sensory readings and detect probes.



\begin{figure}[htpb]
\centering
\includegraphics[width=3in]{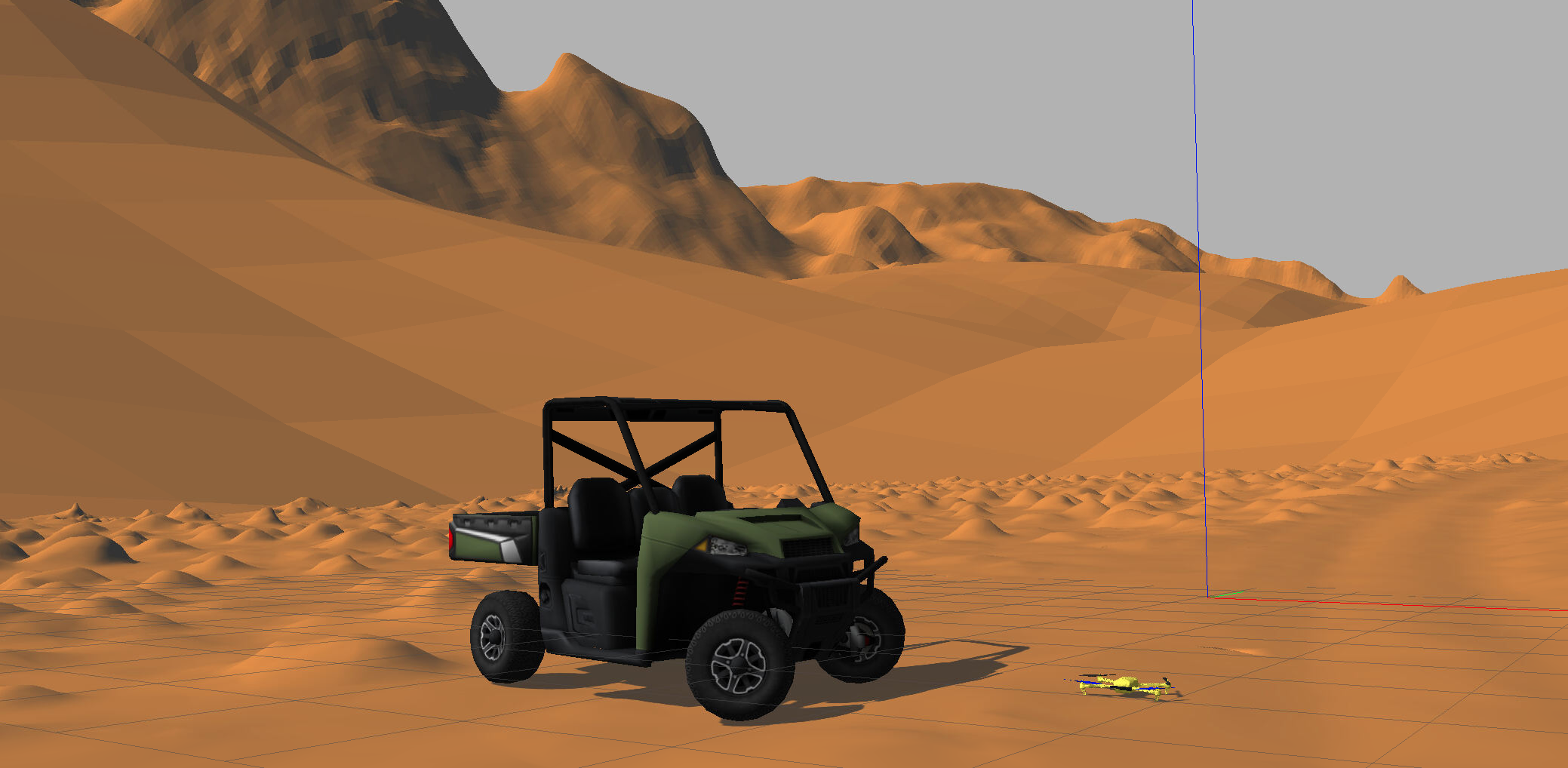}
\caption{\label{fig:environment}{\small}
The simulated environment containing the Mars Jezero world, the rover, and the sUAS.}
\end{figure}

%
Due to the competition requirements, all teams followed the same general guidelines to solve this problem. Each team's methodology begins with a search pattern to find the probe. Each team uses a different object detection algorithm to detect the probe. Once the probe is detected, the sUAS autonomously descend to pick up the probe and it is then carried to a known destination for deployment. Finally, the teams autonomously return the sUAS to the moving rover and land in the trunk. 

\subsection{Team Drone Trekkers (Team 1)}
\begin{figure}[htpb]
\centering
\includegraphics[width=3in]{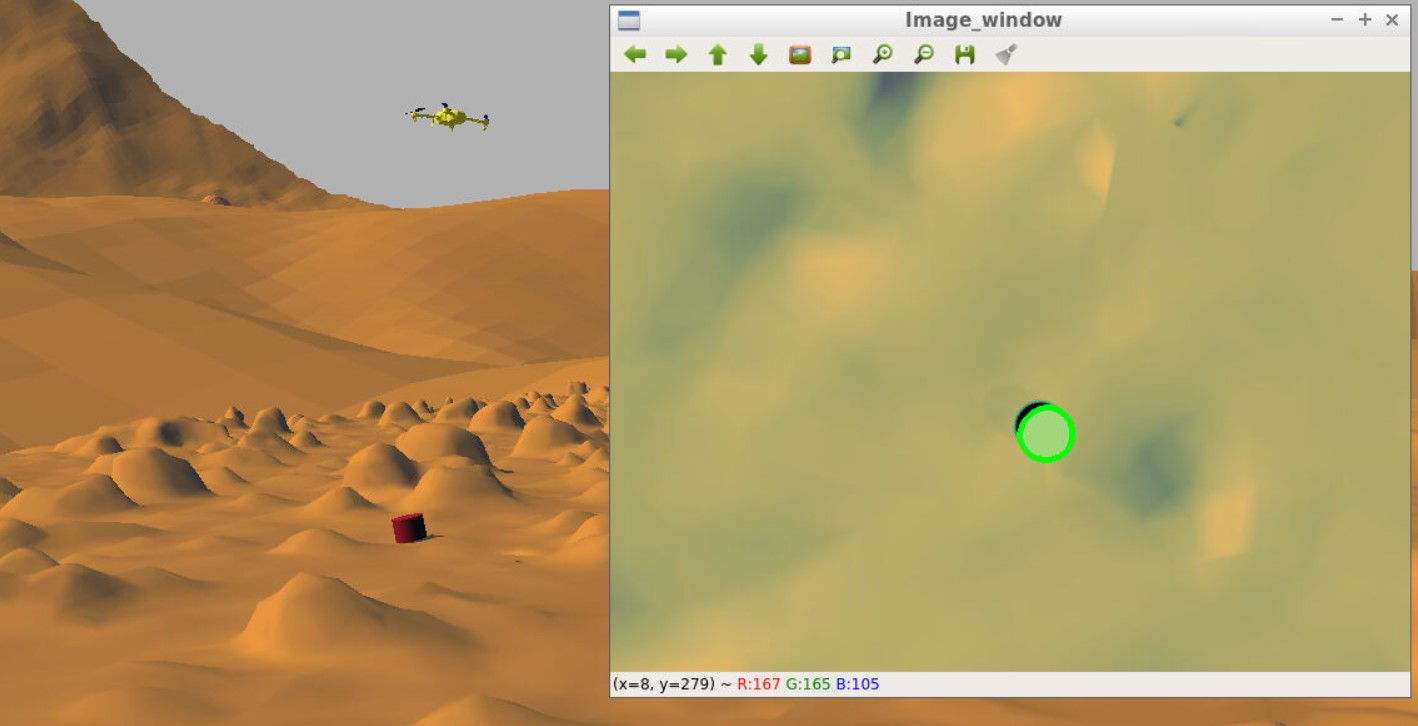}
\caption{\label{fig:Probe_Descent}{\small}Team 1 using color and shape recognition to detect the red probe with their sUAS. The sUAS is in the process of using velocity control to descend onto, and pick up the probe.}
\end{figure}

Team 1's solution for probe detection involves shape and color recognition. The team uses a master controller to switch between sUAS modes to accomplish different tasks (Alg. \ref{alg:alg1}). Terrain relative navigation is implemented using a sonar. The sUAS follows a lawnmower-pattern to search for the probe \cite{williams2001towards}. Then, using a standard digital image processing algorithm of overlaying two HSV images to extract contours, the sUAS detects the probe as a red circle with the largest area. Following that, the sUAS uses velocity control to descend upon the probe, where a Gazebo plug-in attaches the probe model to the sUAS, as seen in Fig. \ref{fig:Probe_Descent}. At this point, the sUAS takes the probe to a predetermined drop location, where it descends and detaches the probe.

\begin{algorithm}[htpb]
\caption{\label{alg:alg1}Team 1's Control Loop}
\While{node OFFBOARDCONTROL has not shut down}{
  \eIf{mode is PROBE}{
   fly$\_$to$\_$destination()\;
   pattern()\;
   hover()\;
   descent()\;
   attach()\;
   mode $\longleftarrow$ \emph{DROP} \;
   }{
   \eIf{mode is DROP}{ 
   fly$\_$to$\_$destination()\;
   pattern()\;
   detach()\;
   mode $\longleftarrow$ \emph{ROVER}\;
   }{
   \eIf{mode is ROVER}{
   \emph{Rover$\_$location} $\longleftarrow$ Kalman$\_$filter()\;
   pattern(\emph{Rover$\_$Location})\;
   detach()\;
   mode $\longleftarrow$ \emph{END}
   }{hover()\;
   continue;}
   }
  }
  Shut down node \emph{OFFBOARDCONTROL}
}
\end{algorithm}

Next, the sUAS navigates to the rover while the rover starts to move forward. The sUAS navigates to the estimated location of the rover. The location estimate is accomplished by combining the GPS location of the rover and the odometry of the rover. The odometry is found by tracking feature points generated using RTAB-map \cite{labbe2019rtab}. This technique refines the location of the rover to better precision. When the sUAS hovers above the rover, it detects the rover using the `You Only Look Once' (YOLO) object detection algorithm implemented through the darknet package \cite{darknet13}. As the sUAS descends, it uses a fiducial marker placed on the rover's trunk to refine the sUAS's trajectory, thus landing successfully into the trunk.

This technique uses the feature tracking odometry from a popular SLAM package, RTAB-map, which uses stereo vision. To estimate the location of the rover, a Kalman Filter is used to combine the odometry-based coordinate tracking and the GPS location of the rover. The filtered location is published as a ROS message that is used by the sUAS to follow and land on the moving rover. A challenge to this approach is that RTAB-map will fail due to a lack of trackable features. This often occurs at places with smoother terrain textures. Due to lack of features, the terrain loses sufficient detail to generate points for RTAB-map. Therefore, when the rover travels to regions with a dearth of features, the Kalman Filter stops working, and the sUAS is unable to follow the rover. However, this problem should not happen in the real world, as there are plenty of features to track in a desert landscape. Therefore, this solution should work in general, but given the online-environment constraints, its success is limited to specific types of terrains.

\subsection{Team Planet Porters (Team 2)}
The algorithm involves the detection and localization of the probe using an image processing technique. The team makes use of a square-spiral search pattern to look for the probe. Once detected, it tracks various red-colored contours using RGB to HSV conversion and tracks the one with maximum area. Once the camera starts providing feedback, a precise set-point control loop is applied to bring the sUAS closer to the probe using the centroid of the contour \cite{fink2015experimental}. The sUAS docks with the probe only when it gets closer than 0.6 meters and when the inclination is less than 30\si{\degree}. The sUAS is in position control mode during this time.

\begin{figure}[h]
\centering
\includegraphics[width=3in]{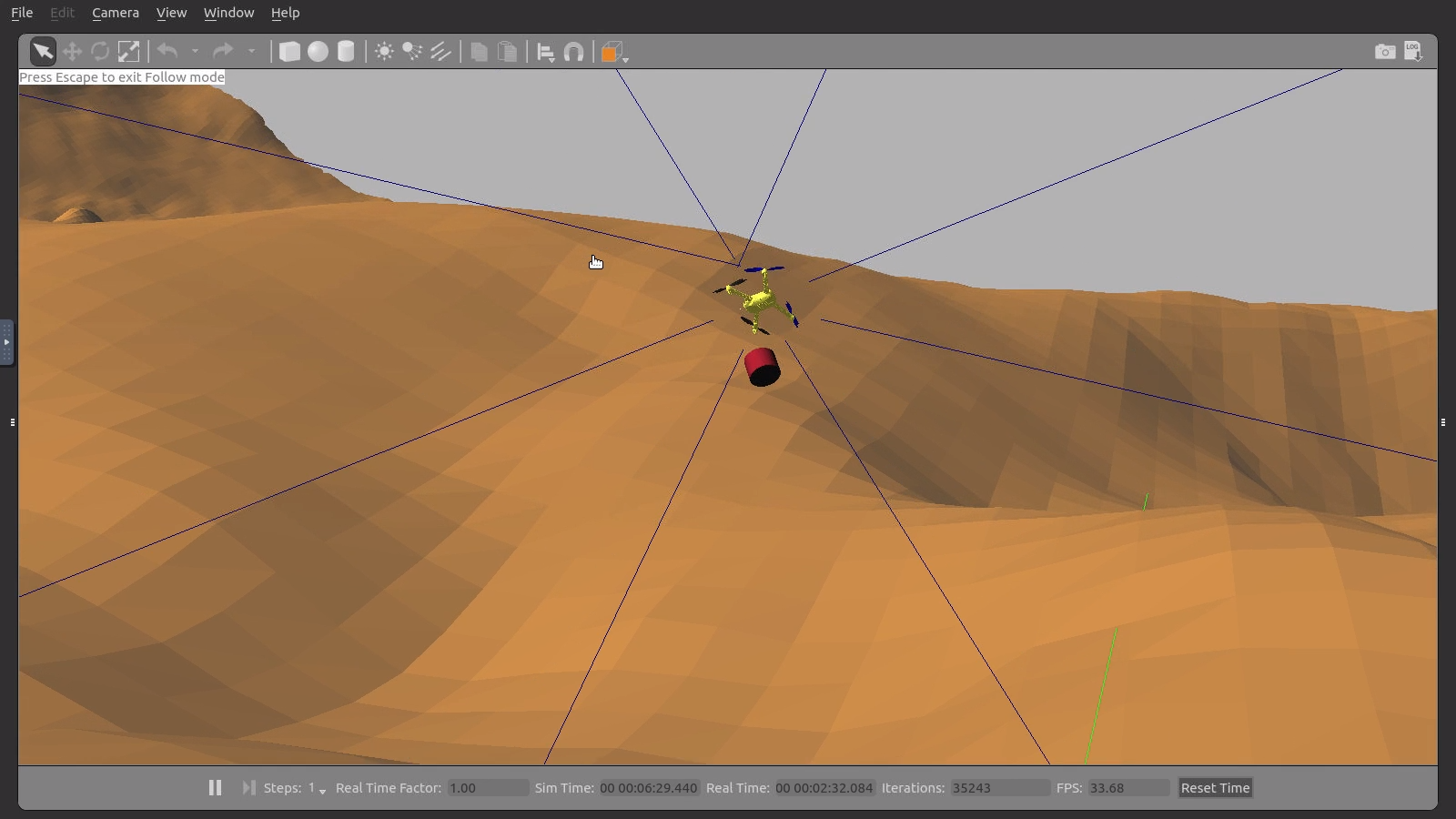}
\caption{\label{fig:rangers} {\small} Team 2 using an array of sonar sensors to detect and avoid the terrain while lifting the sample probe.}
\end{figure}

After the probe is retrieved, the sUAS carries the probe to the drop zone to deploy it. The sUAS advantageously uses the pitch angle during its flight. Eight sonars are installed in different directions on the horizontal plane of the sUAS, which is capable of detecting undulating terrestrial obstacles on its course of flight (Fig. \ref{fig:rangers}). This helps the sUAS detect any upcoming terrain hills easily because of its forward tilt angle. When any sonar senses an obstruction nearby, the sUAS instantly increases its altitude, without altering the horizontal velocity, to avoid collision with the surface. After deploying the probe the sUAS switches to velocity control mode. A velocity limiter is then applied to prevent the sUAS from becoming unstable. 

\begin{algorithm}[hptb]
\caption{\label{alg:pid_alg}Team 2's PID Controller used when the sUAS switches to velocity control mode in order to land on the moving rover.}
time$\_$interval $\longleftarrow$ 0.05;{\COMMENT{specifies the rate,}\\} 
\COMMENT{at which the loop should run}\\
kp $\longleftarrow$ 0.5; 
    \COMMENT{the proportional gain}\\
ki $\longleftarrow$ 0.000005; 
    \COMMENT{the integral gain}\\
kd $\longleftarrow$ 0.4;
    \COMMENT{the differential gain}\\
error$\longleftarrow$ rover$\_$position - sUAS$\_$position; \\
\While{error $\leq$ 0.2}{
previous$\_$error $\longleftarrow$ error; \\
error $\longleftarrow$ rover$\_$position - sUAS$\_$position; \\ 
integral$\_$error $\longleftarrow$ integral$\_$error + error*time$\_$interval; \\ 
differential$\_$error $\longleftarrow$ (error - previous$\_$error)/time$\_$interval; \\
sUAS$\_$velocity $\longleftarrow$ sUAS$\_$velocity + kp*error + ki*integral$\_$error + kd*differential$\_$error; \\
sleep(time$\_$interval); \\
}

\end{algorithm}

After dropping the probe, the sUAS tracks the rover with its GPS coordinates. The solution now applies a PID control algorithm that controls the sUAS's velocity by using the relative displacement between the sUAS and the rover as input (Alg. \ref{alg:pid_alg}). $K_{p}$ is linearly proportional to the system's response. $K_{i}$ eliminates the steady-state error, and $K_{d}$ helps the error converge faster by reducing the overshoot. The rover's trunk is painted red so that the rover's presence can be confirmed visually \cite{falanga2017vision}. The sUAS lowers its altitude linearly as it gets close to the rover and decides to land when it is less than 0.3 meters above the rover's trunk.



\section{Experiments}
\label{sec:experiments}
In this section we, discuss the results of the simulated trials for both methodologies. Each team manually ran 20 trials to determine the efficacy of their approach. Each trial was separated into three steps:

\begin{enumerate}
\item Maneuvering to and retrieving the soil probe. 
\item Deploying the soil probe.
\item Returning to and landing the sUAS in the rover.
\end{enumerate}

To determine the efficacy of the methodologies approaches and to compare the approaches of the teams, two key metrics are defined:

\begin{enumerate}
\item The time taken to complete each step.
\item The consistency between trials, measured by the standard deviation of the sUAS's positions at the same time in different runs of the simulation, and the standard deviation of time to complete the task.
\end{enumerate}

\begin{figure*}[ht!]
  \vspace{6pt}Robit
\centering
\includegraphics[width=6in]{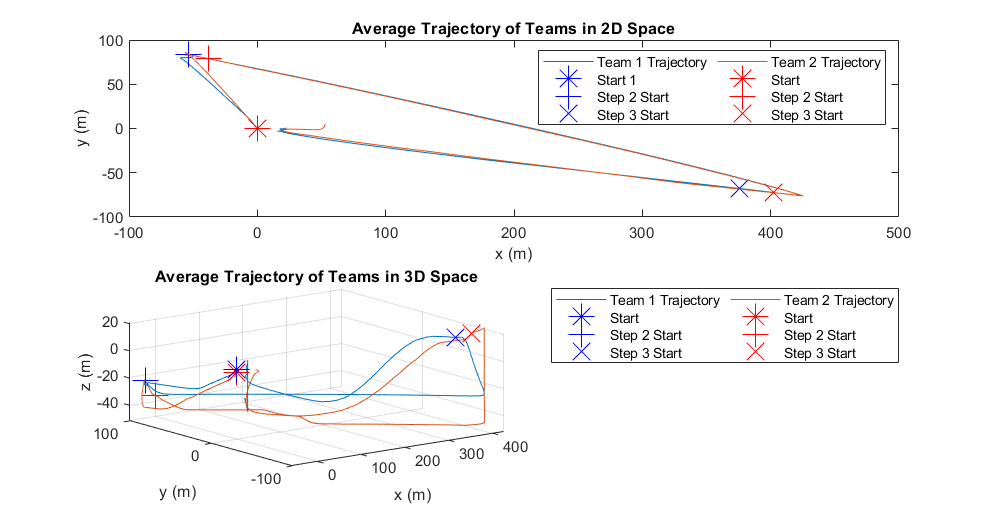}
\caption{\label{fig:Trajectory} {\small}Average trajectories from both teams in 2D and 3D space. Trajectories shown are the average of 20 runs for each team. The location where each step starts is marked. }
\vspace{4pt}
\vspace{-2mm}
\end{figure*}

The graphical representation of these steps can be seen in Figure \ref{fig:Trajectory}. Here, the averaged flight path from the 20 trials is graphed. The start of flight and beginning of each step is indicated for clarity. High levels of consistency between the two teams are observed when the trajectory of the flight is projected in the x-y plane. This is a reasonable observation since the teams are moving to the same positions to search for and deploy the probes. However, there is a significant difference in the flight of the two teams when the sUAS's global z-coordinate, or altitude, is taken into account. This is likely because each team implemented unique and different methods for terrain relative navigation. In addition to this, each team also has a different method for descending onto the rover.

\begin{figure*}[htpb]
  \vspace{6pt}
 \centering
 \includegraphics[width=6.9in]{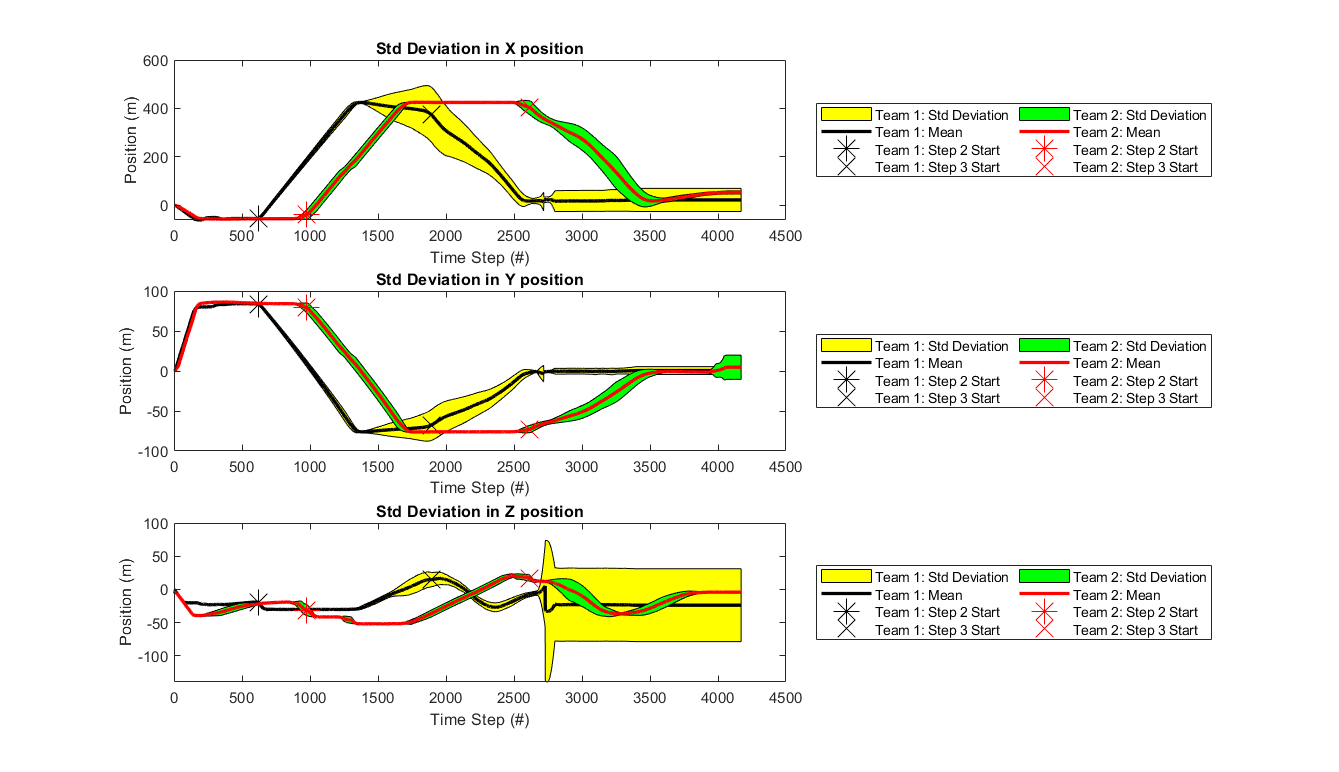}
 \caption{\label{fig:position_deviation}{\small}Standard deviation in position for each axis. The start of each step is marked, and shows Team 1's algorithm completing the steps faster than Team 2. However, this also shows lower deviation in position for Team 2. }
 \vspace{-2mm}
 \end{figure*}

Differences between each team's sUAS-position can be seen explicitly in Figure \ref{fig:position_deviation}. The mean position and the standard deviation of the position are graphed for each team. Through these graphs, we can determine that Team 2 built a more consistent algorithm with respect to the position of the flight at the same time for each simulation. During the first step, the two teams remained highly consistent. This is a logical result considering that the sUAS started from the same location and looked for the probe, which was always in the same location. It is not until the beginning of the third step that we start to see major delineations in the trajectory of flight between the two teams.

The standard deviation of the trajectory of flight for both teams increases in step 3. 
This is a reasonable observation since, at this stage, the sUAS is required to land on a moving rover. This part of the mission adds uncertainty to the flight. In this scenario, the second team's flight has a  
lower standard deviation. The second team's flight is more likely to end in the same location through each episode, thus decreasing their final standard deviation. The difference between the team's trajectory of flight is most prominent in the z-direction.

In addition to this, Figure \ref{fig:position_deviation} also highlights key information about the time elapsed in the flight between teams. Although Team 2 is more consistent, Team 1 performs the steps much faster. If the teams start at the same time, Team 1 moves to Step 2 faster than Team 2. These differences are tabulated in Table \ref{tab:table1}.

\begin{center}
\begin{table}[h!]
    \centering
    \begin{tabular}{||c|c|c|c|c||}
    \hline
     & \multicolumn{2}{c|}{Mean Time (s)} & \multicolumn{2}{c||}{Standard Deviation (s)} \\
    \hline
     Step & Team 1 & Team 2 & Team 1 & Team 2 \\ [0.5ex]
     \hline\hline
     1 & 50.05 & 93.34 & 0.971 & 4.154 \\
     \hline
     2 & 97.70 & 163.56 & 12.01 & 2.253 \\
     \hline
     3 & 125.59 & 228.85 & 12.76 & 12.20 \\
     \hline
\end{tabular}
\caption{\label{tab:table1}Time comparisons between each team, separated for each step of the run.}
\vspace{-8mm}
\end{table}
\end{center}

It is observed that Team 1 completes each step faster. The average time to complete the entire trial for Team 1 was 2742.0 seconds, while it took 3942.0 seconds for Team2. Furthermore, the teams are able to run these episodes at a similar level of consistency. Team 1 completes Step 1 with a much lower standard deviation than Team 2. On the other hand, Team 2 completes step 2 more consistently. Finally, the standard deviation for step three is similar for both teams. Therefore, Team 1 will always be faster than Team 2, and both teams are equally consistent in the time domain.

Based on the metrics defined above, Team 2 follows a better methodology. This is due to the overall consistency of the team's method. Even though Team 1 completed their tasks much faster, Team 2 could do so more reliably. Given the Mars 2020 use-case that drives these methods' development, reliability is a more important metric than the time taken.


\section{Discussion}
\label{sec:discussion}
Under normal circumstances, this competition would have taken place outdoors, and software would have been tested on flyable hardware. This was not possible due to the COVID-19 pandemic. However, the OpenUAV platform enabled this competition to continue in a meaningful way.

As shown above, both teams were able to implement a complex methodology to approach the problem. The simulation framework allowed for intricate autonomy solutions while allowing other, less experienced competitors to learn the basics of the ROS and Gazebo framework. Therefore, continuing education on Cyber-Physical Systems at all skill levels throughout the pandemic. Moreover, the differences in the team's approaches highlight the versatility of the container environment. Practically, what can be implemented and tested in real-life can also be implemented on the container.

This is true despite the challenges faced while using RTAB-Map. Feature tracking is possible in regions of the simulation where the terrain had detailed features. In other regions, where the terrain lacks useful features, tracking becomes challenging. Hence, the problem is with the simulated environment, not with the OpenUAV platform.

Furthermore, the OpenUAV platform enabled certain benefits that would not have been possible under normal circumstances. For example, the winning team competed from India, whereas Team 1 competed from the United States. This collaboration would not have been normally possible for logistical reasons. Plus, the platform enabled team members to work together on the same container from different locations. This created a comfortable environment for mentorship, where more experienced team members could easily walk through complex code and methodologies with less experienced members. 

Competition participants were also all enthusiastic about participating in the competition. Most internship, REU opportunities, and other in-person sUAS competitions were canceled due to the pandemic. Using the OpenUAV platform, it was trivial to modify the original CPS challenge's scope to move online. Thus, providing students a place to continue to learn applied skills in a structured environment. This level of resilience allowed for meaningful work and continued education to take place while most other programs had to be stopped.

The most considerable downside to an online environment is that the teams could not build physical sUASs and fly them to test their algorithms. There is much to be learned by comparing simulation results to a real application, especially for aspiring engineers. However, by removing the competition's hardware aspect, the teams had a more in-depth and thorough experience with the software side. This helped develop valuable software skills in students who would not have been exposed to them otherwise. Now, when the pandemic is over, competitors will be better overall engineers because they will know how to program their physical systems. Plus, lessons learned during this competition about robotic control will inform future physical designs.

\section{Conclusions and Future Work}
\label{sec:future-work}

In this paper, we presented the methodologies and results of a virtual robotics challenge. We demonstrated how competitions such as these could effectively continue in a virtual environment. Furthermore, we have demonstrated how education in Cyber-Physical Systems can continue virtually. Tools addressed in this paper can continue to be used for simulation and extended collaboration even after the pandemic. Containerization and OpenUAV can also significantly lower the barrier for entry into robotics, as students do not need to have local access to the computing infrastructure necessary to run complex simulations.

As part of the NSF-sponsored competition, we performed and analyzed two potential methodologies for probe recovery on Mars. Given the goals of Mars 2020, these methods can be an essential proxy for actual mission settings. Moreover, with the correct soil probe, these algorithms are directly applicable to Earth and can be reused for terrestrial exploration associated with geological and ecological studies.

Future work can branch in two different directions. The next step is to build sUASs and test the algorithms as initially intended in the hardware path. The tested algorithms can then be directly used to deploy and retrieve soil probes on Earth. In the software path, the simulation and platform can continue to be refined. Now that Mars 2020 has landed on Mars, more data and scans of the simulated environment will become available. Thus, the simulated environment can continue to be enhanced, and more accurate algorithms can be developed.

{\small \section{Acknowledgements}
\label{sec:acknowledgements}
We gratefully acknowledge NSF grant CNS-1521617 for this work.}
{\small
\printbibliography}

\end{document}